\documentclass[letterpaper, 10 pt, conference]{ieeeconf}  

\IEEEoverridecommandlockouts                              

\usepackage{graphicx} 
\usepackage{booktabs}
\usepackage{multirow}
\usepackage{amsmath}

\usepackage{subfigure}
\usepackage{adjustbox}

\usepackage{xcolor} 

\title{\LARGE \bf
TK-Planes: Tiered K-Planes with High Dimensional Feature Vectors for Dynamic UAV-based Scenes
}

\author{Christopher Maxey\textsuperscript{\rm 1,2}, Jaehoon Choi\textsuperscript{\rm 2}, Yonghan Lee\textsuperscript{\rm 2},  Hyungtae Lee\textsuperscript{\rm 3}, Dinesh Manocha\textsuperscript{\rm 2}, and Heesung Kwon\textsuperscript{\rm 1}\\
\textsuperscript{\rm 1}DEVCOM Army Research Laboratory \textsuperscript{\rm 2}University of Maryland at College Park \textsuperscript{\rm 3}BlueHalo
\thanks{Correspondence to cmaxey@umd.edu}%
}

\begin{document}

\maketitle
\thispagestyle{empty}
\pagestyle{empty}

In this paper, we present a new approach to bridge the domain gap between synthetic and real-world data for unmanned aerial vehicle (UAV)-based perception. 
Our formulation is designed for dynamic scenes, consisting of small moving objects or human  actions. 
We propose an extension of K-Planes Neural Radiance Field (NeRF), wherein our algorithm stores a set of tiered feature vectors. The tiered feature vectors are generated to effectively model conceptual information about a scene as well as an image decoder that transforms output feature maps into RGB images.  Our technique leverages the information amongst both static and dynamic objects within a scene and is able to capture salient scene attributes of high altitude videos.  We evaluate its performance on challenging datasets, including Okutama Action and UG2, and observe considerable improvement in accuracy over state of the  art neural rendering methods.

\section{Introduction}

Synthetic data has been used for over a decade to train deep learning algorithms in a variety of tasks, from game play \cite{mnih2013playing} to driving \cite{song2023synthetic}. Data from sources such as game engines \cite{plowman20163d,shah2018airsim,alvey2021simulated} and generative algorithms \cite{goodfellow2020generative,rombach2022high} have made significant advancements in recent years, particularly in enhancing fidelity and bridging the domain gap between synthetic \cite{shen2023progressive,shen2023archangel} and real data \cite{maxey2023uav}.
Generative algorithms for instance can create realistic novel imagery and videos from a given prompt \cite{rombach2022high,Sora}, requiring an implicit model of real world objects and their codependencies.
Neural Radiance Fields \cite{mildenhall2021nerf} and, more recently, Gaussian Splatting \cite{kerbl20233d} have demonstrated remarkable capabilities in rendering both specific 3D static and dynamic scenes, providing novel views of a scene that were not available during the training process.
Because of these advances, synthetic data have been playing a larger role in supplementing existing real world datasets for training data hungry perception algorithms \cite{redmon2016you,feichtenhofer2020x3d}. 

\begin{figure}[t]
    \centering
    \includegraphics[width=1.0\linewidth]{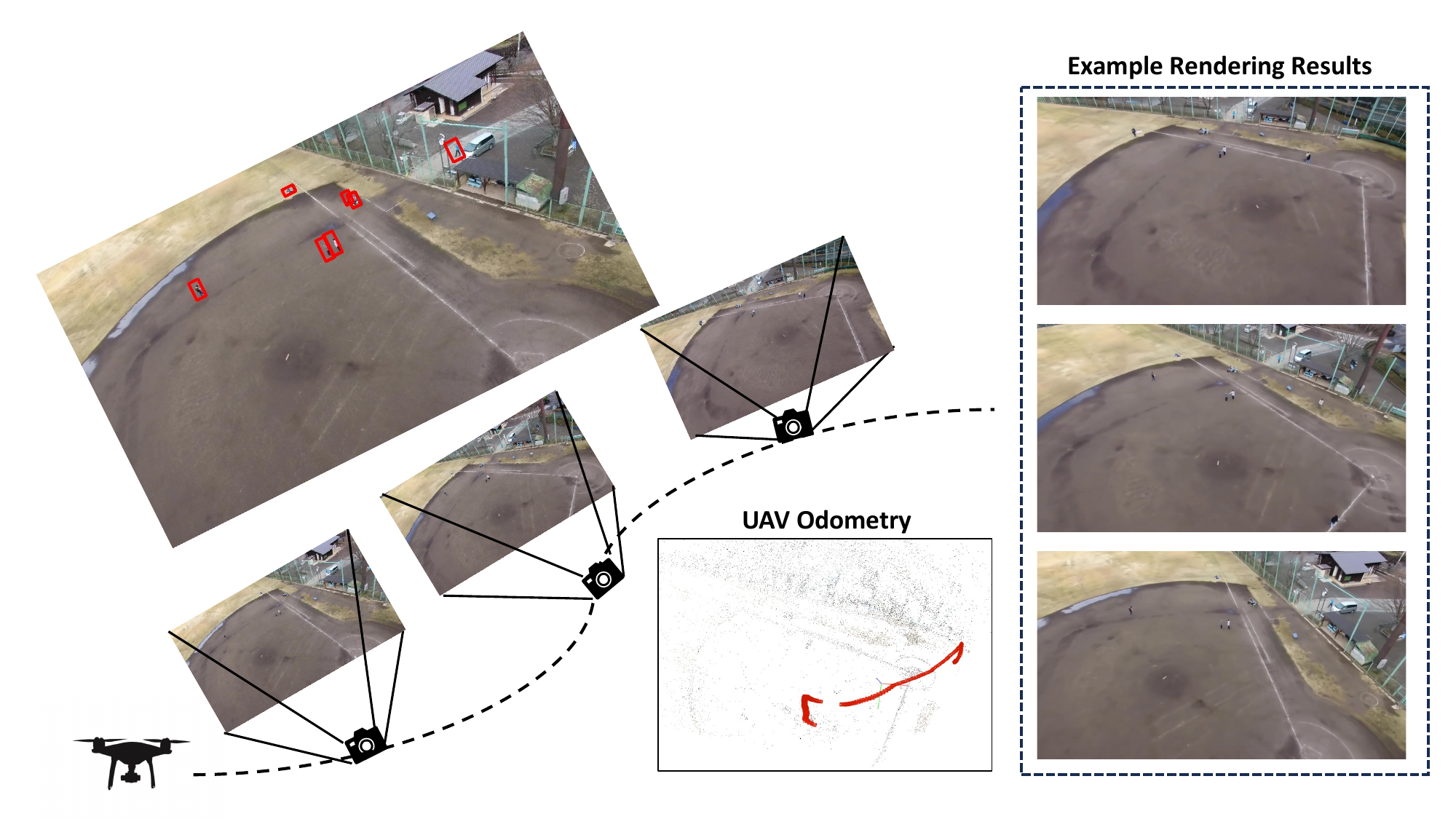}
    \vspace{-5mm}
    \caption{Example neural rendering results. Our target dynamic scene contains small, fast-moving people (in red box) captured using only a monocular camera. Despite this challenging scenario, our proposed method produces high-quality renderings for dynamic scene.} 
    \vspace{-5mm}
    \label{fig:intro}
\end{figure}

Specific domains, such as videos captured by UAVs \cite{barekatain2017okutama,li2021uav,shen2023archangel}, present unique challenges stemming from environmental conditions and a relative scarcity of datasets within these domains. In regard to UAVs, perception is not limited to static tasks such as object detection within a single frame but can also include more complex tasks like action recognition which can require multiple frames of a dynamic agent.  The inclusion of dynamic objects and agents further compounds the difficulty of a task and accentuates the need for large amounts of data.  Esoteric tasks, for example search and rescue, add another layer of complication as more general datasets may not capture the requisite environmental detail and conditions or relevant agent behavior.

Given a dynamic UAV scene, several issues become pertinent when trying to model accurate dynamic behavior.  First, because of the nature of high altitude data, target objects are often at a large distance from the camera and appear small relative to the background.  Due to the more difficult task of modeling dynamic objects, this imbalance between static and dynamic elements of a scene can cause issues with separating static and dynamic elements within any given algorithm, in particular if there are inaccuracies present in camera pose information.  Second, dynamic objects are often sparse in the scene at any given point of time.  This further limits the amount of data that can afford training.  Lastly, diverse poses at specific time stamps may be limited, often with just one camera pose per timestamp and limited variation in camera poses.  This can lead to degeneracies in the output of a model the more poses and their corresponding timestamps diverge from the training data.

\noindent {\bf Main Results:} Our method aims to rectify these issues with a novel Neural Radiance Field architecture \cite{mildenhall2021nerf}. We present a new approach to learn NeRF-based models for UAV-perception, that can create novel-view images from previously unseen camera positions capturing
salient attributes of the scene. Our approach is used to augment current datasets by generating synthetic images for improved pose and action recognition. 

We present a tiered version of the K-Planes algorithm \cite{fridovich2023k}, TK-Planes, that outputs and operates on feature vectors rather than RGB pixel values.  These feature vectors can store conceptual information about a particular object or location within a scene and  when output for multiple corresponding camera rays, form feature maps that are decoded to form a final image.  Furthermore, taking advantage of the fact that many details of a scene, including dynamic objects, repeat throughout, the image decoder which operates on all feature vectors can learn how to model static and dynamic objects within the scene.

We have evaluated the performance on some challenging UAV datasets, including Okutama Action~\cite{barekatain2017okutama} and UG2~\cite{vidal2018ug}, which consist of high altitude videos. We show that TK-planes can generate high fidelity imagery of UAV scenes, capturing small dynamic details well and outperforming previous state of the art neural rendering methods.

\begin{itemize}
    \item Development of a novel grid-based NeRF architecture that utilizes multiple scales of feature space in order to render frames from a dynamic UAV scene.
    \item Extension of previous pipelines for training grid-based NeRF architectures to include an image decoder.
    \item Showcasing high fidelity renderings with PSNR and dynamic-PSNR (DPSNR) values exceeding previous methods such as K-Planes \cite{fridovich2023k},  Extended K-Planes \cite{maxey2023uav} and 4D Gaussian Splatting \cite{wu20244d}.
\end{itemize}


\section{Related Works}
\noindent\textbf{Synthetic Data Generation.} Collecting real-world dynamic dataset can be expensive, especially for challenging environments such as unmanned aerial vehicle (UAV)-based scenes. One way of generating synthetic dataset is utilization of game engines or simulators \cite{plowman2016game, shah2018airsim, alvey2021simulated}. While these rendering pipelines can generate images in a fully controlled environment setting, they require large amount of efforts and skills for setup. Meanwhile, recent advances in neural rendering \cite{xie2022neural,chen2024survey} show surprising performance in novel view synthesis, which shows their feasibility to augment sparse image datasets with arbitrary rendering in a novel view \cite{maxey2023uav,ge2022neural}  and appearance \cite{Yang_2023_CVPR,wang2023neural}, or even with injected objects in the scene \cite{wang2023neural}. 

\noindent\textbf{Neural Rendering for Dynamic Scenes.} To represent 3D dynamic scenes, various NeRF research explores 4D neural radiance fields representation with an additional temporal dimension \cite{du2021nerflow, xian2021videonerf, gao2021dnvs, wang2021dctnerf}. Instead of embedding continuous time dimension into the radiance fields, some of variants uses time-varying latent code \cite{li2022dynerf, peng2023mlpmap}. While these 4D NeRF show capability to represent complicated dynamic scenes, they generally suffer from inherent representational redundancy when they learn 4D contents from 2D imagery only. The other major variants of NeRF incorporates canonical scene and a deformation field \cite{park2021nerfies, pumarola2020dnerf}. However, these methods demonstrate limited capability to learn topological changes or large motion in the scene due to their heavy dependence on the canonical scene.  Recently, Gaussian Splatting \cite{kerbl20233d} has become a state of the art research topic for neural rendering, producing accurate results with fast rendering speeds. \cite{wu20244d} applies Gaussian Splatting to temporal scenes, outperforming previous 4D neural rendering methods in both speed and accuracy.

\noindent\textbf{NeRF with Explicit Grid Representation} 
Integration of 4D NeRF models with explicit voxel grid structures \cite{guo2022ndvg, liu2022devrf, peng2023mlpmap, fang2022tineuvox} has been widely studied to boost slow training and rendering speed of implicit NeRF models. However, voxel-based methods have limitation in representing large-scale scenes due to their inefficient use of memory storage. One way to achieve compact explicit representation is factoring 4D dynamic volumes into multiple planar features planes or volumes \cite{fridovich2023k, cao2023hexplane, shao2023tensor4d, gao2023strivec, mihajlovic2024resfields, wu2024tetrirf}. These tensor factorization techniques show good trade-off between compactness and high-quality representation. However, all of these methods primarily target dynamic scenes observed from ground-level perspectives, typically featuring a single person exhibiting transient motions close to the camera. UAV-Sim \cite{maxey2023uav} shares a similar motivation as ours, utilizing extended K-planes for dynamic scene representation and addressing multiple human actors captured by UAVs. We enhance their method \cite{maxey2023uav} by introducing higher-level features to augment the low-level detail of multiple human actors.

\section{Methodology}

\subsection{Novel View Synthesis for Dynamic Scenes}

Given a difficult domain such as UAV video with limited available training data and a small dynamic to static pixel ratio, our approach utilizes NeRF in feature space in order to better capture and render objects of interest within a scene, e.g. dynamic objects.  We employ a grid-based Neural Radiance Field, the grids of which are similar to those from K-Planes \cite{fridovich2023k}.  A key difference being that the grids in our algorithm store feature vectors that do not directly encode for RGB values but rather for more abstract high level concepts within the scene.  As such, our NeRF algorithm outputs a series of feature vectors and density values for a given ray which are then combined via typical volumetric rendering.  These combined feature vectors form a feature map as opposed to an RGB image.  The output feature maps are decoded by a convolutional neural network in order to generate a final rendered RGB image.

\subsection{Neural Radiance Field Overview}

To understand our algorithm in more detail, a brief overview of Neural Radiance Fields follows.  Neural radiance fields are a type of volumetric rendering that utilize neural networks to output a color value and density given an input camera pose.  From a designated camera pose, rays are cast into a scene and incremental points along each ray are sampled.  Typically, two Multi-layer Perceptrons (MLPs) are employed, one for spatial density and another for a directionally dependent RGB value.  The spatial density MLP takes as input a position within 3D space and outputs a density value as well as a feature vector.  This feature vector and the given ray direction are fed into the second MLP which outputs a RGB color value.  The density and color value at each sampled point along the ray (Eq. \ref{eq:combined}-(1)) are then accumulated via volumetric rendering.  Eq. \ref{eq:combined}-(2) depicts the overall color value given a ray wherein the point specific color value, density, and transmittance (Eq. \ref{eq:combined})-(3) up to the given point are multiplied and integrated.  The transmittance is a representation of how much light will pass through all previous points and thus how much the current point contributes to the overall color value.  This version of NeRF relies on the MLPs learning an implicit 3D structure of a scene.




\begin{align}
r(t) &= o + td \\
C(r) &= \int_{t_n}^{t_f}T(t)\sigma(\mathbf{r}(t))c(\mathbf{r}(t), d)dt \\
T(t) &= exp(-\int_{t_n}^{t}\sigma(\mathbf{r}(s))ds)
\label{eq:combined}
\end{align}

Alternative to an implicit method, some NeRF algorithms implement an explicit representation of a scene stored in a data structure such as a matrix \cite{guo2022neural, sun2022direct} or a series of grids \cite{fridovich2023k, cao2023hexplane}. These data structures contain feature vectors that learn details of a scene and when passed through a density and color MLP, output density and color values respectively. The explicit spatial, and temporal if applicable, layout of the data structures offloads the necessity for implicitly learning the spatial and temporal structure of the scene within the MLPs. The benefit of this is that smaller MLPs can be used resulting in quicker training and inference times.  The downside is increased memory usage, in particular for matrix representations, though the use of factored grid representations \cite{fridovich2023k, cao2023hexplane} helps to alleviate this requirement.


\subsection{Tiered Planes in Feature Space}

\begin{figure}[t]
\centering
\includegraphics[width=0.90\linewidth]{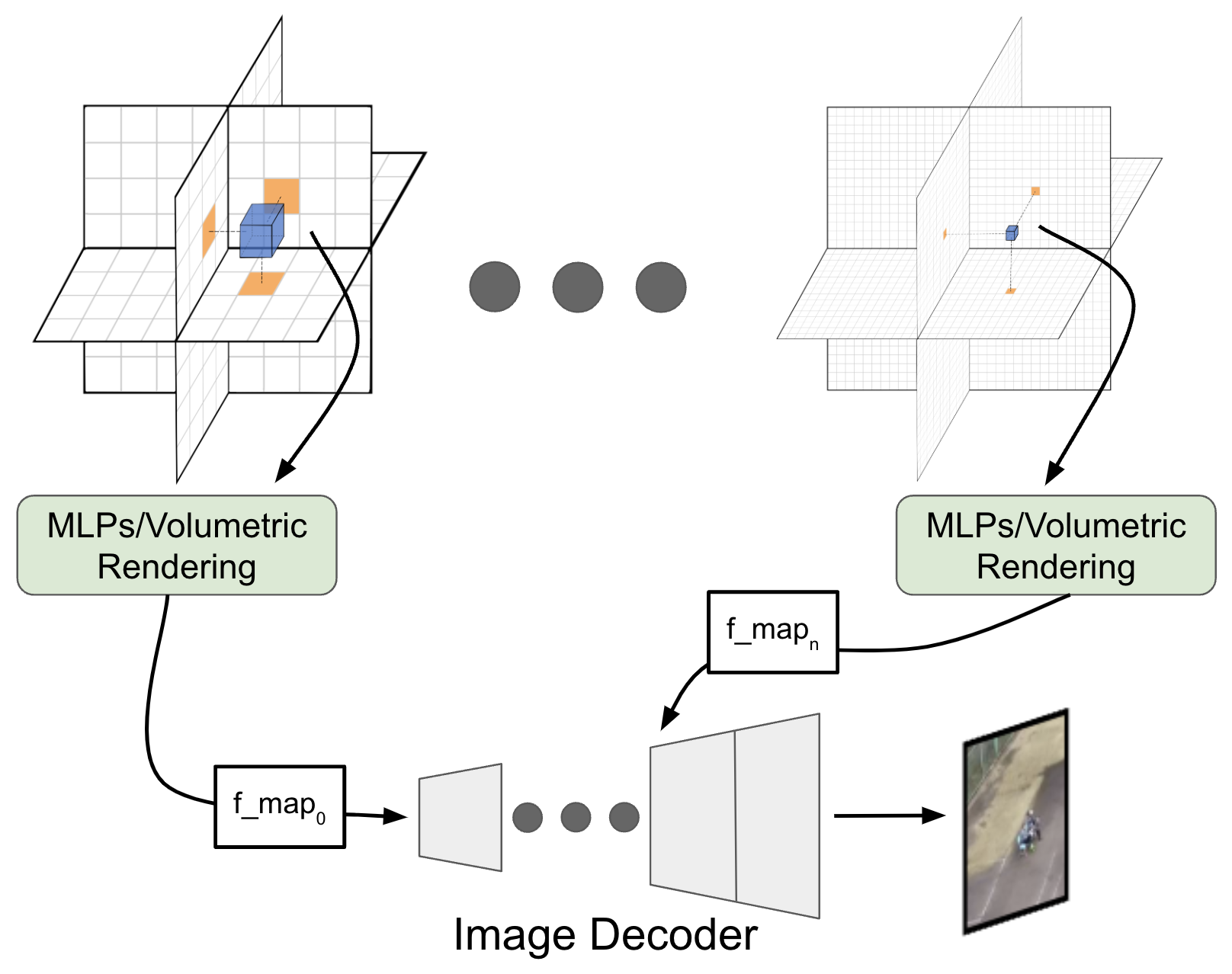}
\vspace{-1mm}
\caption{Overview of our tiered planes algorithm:  An arbitrary $n$ number of grids are shown, representing different scales of feature space, with larger scales (earlier feature maps) capturing more abstract scene details.  Note that the algorithm uses nine grids in total per scale, three static spatial, three dynamic spatial and three dynamic spatio-temporal.  Feature vectors are processed via concurrent MLPs to output a density value and final feature vector.  These are accumulated with volumetric rendering into a feature map.  Feature maps from each set of grids are input into corresponding stages of the image decoder.}
\label{full_system}
\vspace{-6mm}
\end{figure}

Our main approach operates in a conceptual feature space rather than an RGB space with respect to the output of the NeRF algorithm.  The main purpose of focusing on feature space rather than RGB space is to learn feature vectors that represent cohesive objects of a scene, such as people, as opposed to learning solely a color and density value.  This helps to alleviate the issue of poor fidelity of high interest objects across different poses and time-slices.  We use the same grid representation as seen by K-Planes \cite{fridovich2023k} with changes influenced from the extended version as seen in \cite{maxey2023uav}.  Figure \ref{full_system} shows our system in full which includes abstract feature grids (up to $n$) as well as an image decoder. Our NeRF model outputs a feature map rather than a rgb image and the image decoder processes the feature map into a final image.

An ancillary benefit of TK-Planes is a vast decrease in training and inference times when compared to previous state of the art NeRF methods.  In general, explicit factored representations of a scene improve the speed of training and inference; however, a bottleneck still exists due to the ray casting and ray sampling required for NeRF techniques, wherein a large number of poses (on the order of 10's of thousands) within a scene are processed. A reduction in the number of points required during sampling has a direct impact on the speed of training and inference.  Our method allows for a considerable reduction in the number of rays needed to render an image as well as the number of points along each ray that need to be sampled.  This is due to the fact that the output of our NeRF model is a feature map rather than a RGB image and said feature maps can have height and width dimensions much smaller than that of the final image.  For instance, the final image dimensions in the Okutama-Action dataset are 720x1280 yet the feature map dimensions used in our experiments for this dataset start at 45x80, an overall reduction factor of 256 with respect to the number of rays needed to render a single image.

\subsection{TK-Planes Architecture Details}

As seen in \cite{maxey2023uav}, we add the additional planes for spatial features that are dynamic. The resulting static and dynamic features are described as follows:

\begin{equation} \label{eq:2a}
f_{s}(\mathbf{q}) = \prod_{c \in C_{s}}f_{s}(\mathbf{q})_{c}, \;\;\;\;f_{d}(\mathbf{q}) = \prod_{c \in C_{d}}f_{d}(\mathbf{q})_{c}.
\end{equation}

\begin{equation} \label{eq:2b}
f(\mathbf{q}) = f_{s}(\mathbf{q})\oplus f_{d}(\mathbf{q})
\end{equation}

wherein $f_{s}$ is a function of the static spatial planes and $f_{d}$ is a function of the dynamic spatial and temporal planes and $\prod$ represents the Hadamard product.  In our version, each feature vector $f_{*}(\mathbf{q})_{c}$ is a $D$ dimensional vector wherein $D$ is a hyperparameter.  From \ref{eq:2b}, a grouping of feature vectors $f(\mathbf{q})$ is calculated by concatenating $f_{s}$ and $f_{d}$ in the batch dimension such that each spatial and dynamic feature vector can be processed independently.  $f(\mathbf{q})$ is then input into the MLP networks to output a density value and another feature vector for each static and dynamic feature vector.  Volumetric rendering along the ray returns a final spatial and dynamic feature vector value per pixel of an output feature map at a given feature space scale, $d_{s}$.  To maintain a balance between speed, accuracy and memory usage, our experiments include using up to two feature space tiers which allows for training using an entire image rather than an image patch.

K-Planes \cite{fridovich2023k} and its extended version \cite{maxey2023uav} use multi resolution grids offering different levels of scene details.  \cite{wu20244d} also uses a k-planes style multi-resolution grid system to sample small feature vectors for processing gaussian parameters. The resulting feature vector from each level can either be summed or concatenated before being processed by the MLPs and in the case of NeRF models, then accumulated via volumetric rendering.  In these cases, the volumetric rendering is the final step.  However, in our algorithm it is more intuitive to separate the different levels of detail as input into appropriate stages of the image decoder rather than as input entirely into the front of the decoder.  Convolutional image decoders operate on high-level abstract features to low-level detail features as their stages progress and typically benefit from feature map input at varying stages \cite{ronneberger2015u}.  This allows the grids at varying scales of feature space to learn appropriate details with respect to rendering an output image.


\begin{figure}[t]
     \centering
     \subfigure[Main Image Decoder Block]{
         \includegraphics[width=0.45\textwidth]{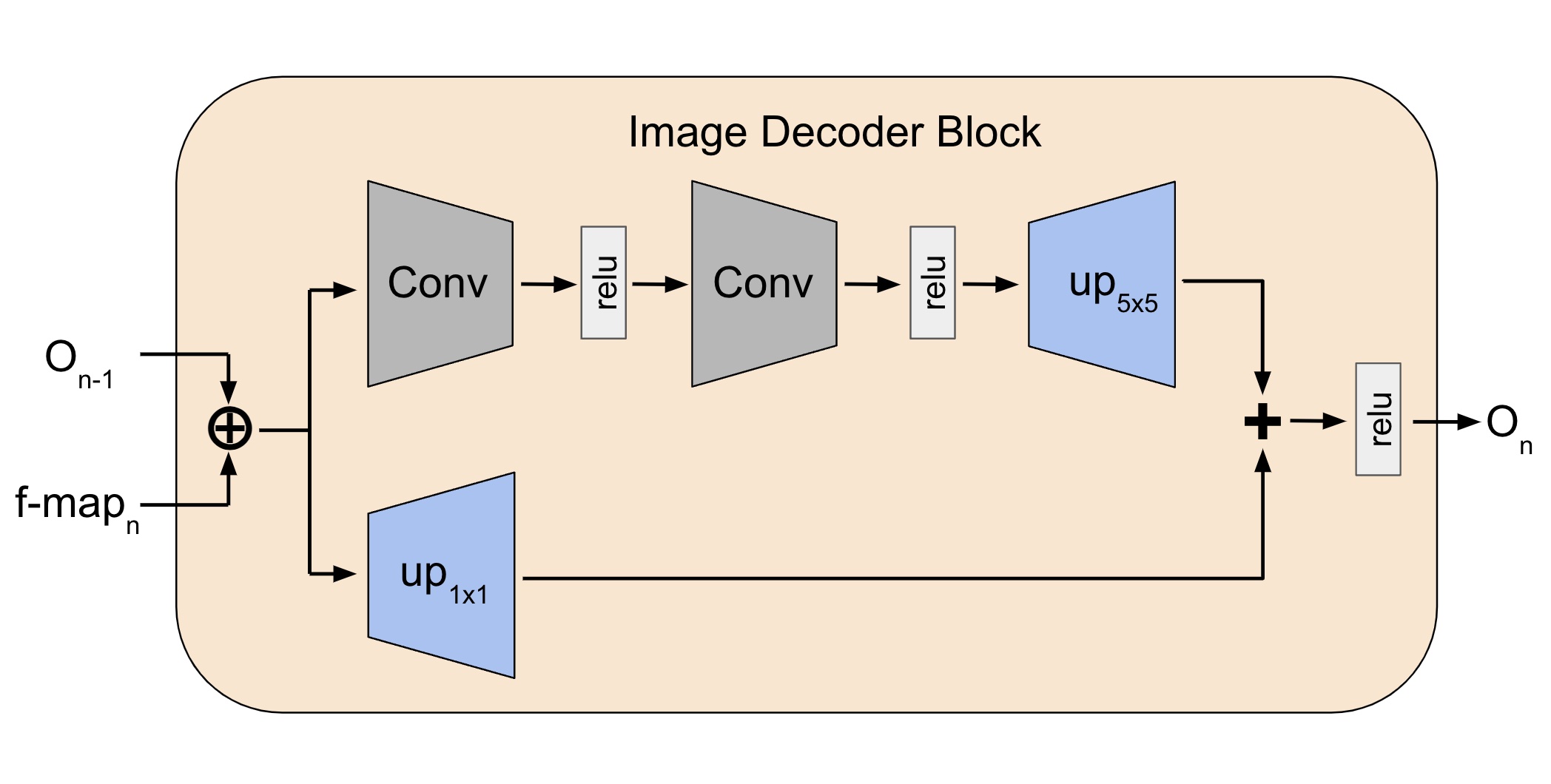}
        \label{fig:compare_okutama}
    }
     
     \subfigure[Final Image Decoder Block]{
    \includegraphics[width=0.46\textwidth]{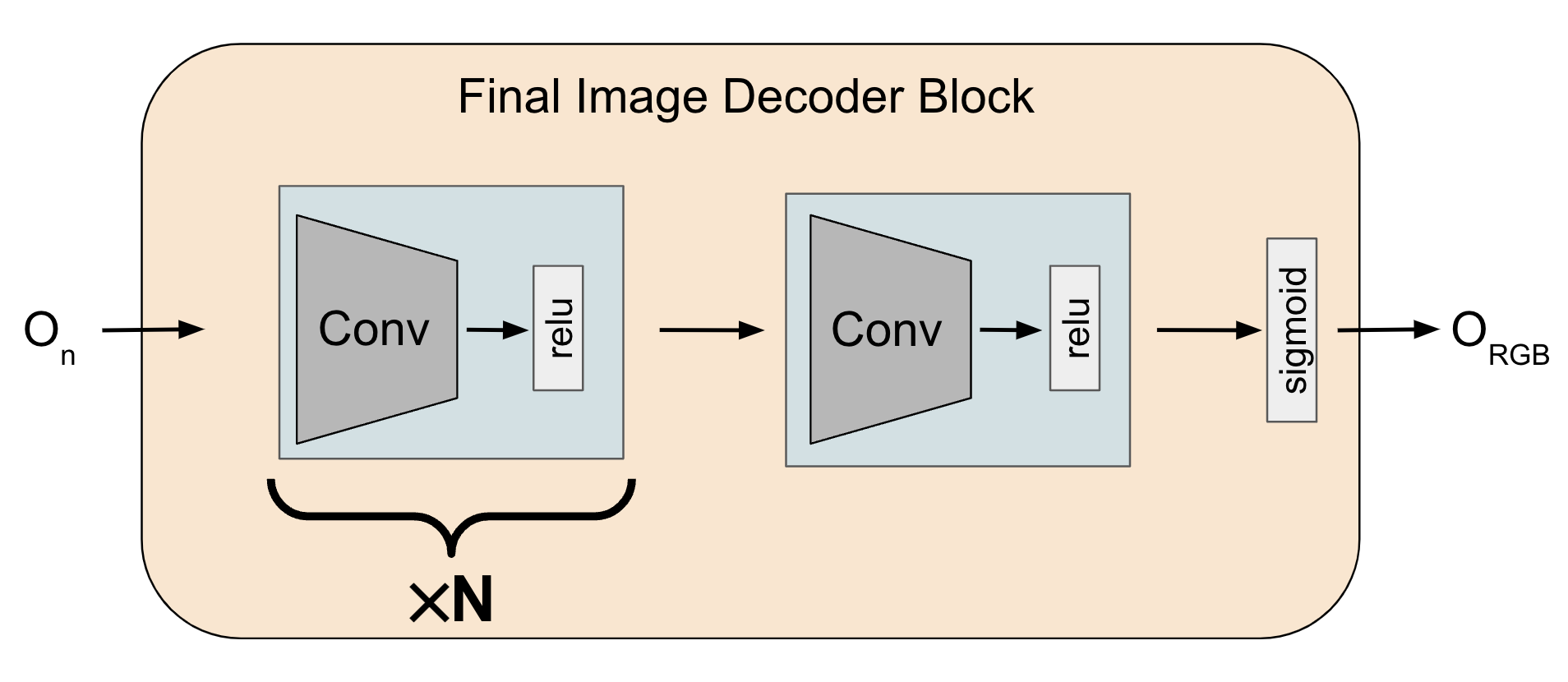}
        \label{fig:compare_kplanes}
     }
     \caption{a.) A diagram of a block within the image decoder.  The $n^{th}$ block accepts up to two inputs, a feature map from the $n^{th}$ set of grids, if applicable, and the output $O_{n-1}$ from the previous block.  At $n=0$, there is no input from a previous block, only the feature map.  Beyond the last set of grids, there is no feature map input, only the input from the previous block. b.) The final image decoder block with $\text{x}N$ $3\text{x}3$ convolutional layers followed by a final convolutional layer to reduce the channel size to three and a sigmoid activation layer to rectify the raw output into RGB values.}
\label{decoder_blocks}
\end{figure}

\subsection{Image Decoder}

The image decoder is a multi-stage process wherein each stage consists of two upsampling layers, one for a skip connection ($up_{1\text{x}1}$) and another for upsampling the processed input ($up_{5\text{x}5})$, and a processing block consisting of two convolutional layers.  $up_{1\text{x}1}$ consists of a bilinear upsampling that doubles each spatial dimension of the input, followed by a $1\text{x}1$ convolutional layer that halves the channel dimension of the input.  As mentioned, this serves as a skip connection between the input and processed feature maps and is meant to minimally affect the input value.  $up_{5\text{x}5}$ consists of a bilinear upsampling that doubles each spatial dimension of the processed input and a $5\text{x}5$ convolutional layer that reduces the channel dimension of the processed input to half of the original input channel dimension.  A $5\text{x}5$ kernel size is chosen for reducing the channel dimensions since the processed input spatial dimensions are doubled, thus a $5\text{x}5$ kernel offers a larger receptive field of the initial input.  For each stage, the static and dynamic feature maps are processed independently using the same layers for each.  This was chosen to avoid additional complexity in the decoder and to ideally contribute to layer robustness. 

After all upsampling stages, a final stage combines the static and dynamic feature maps into an output image.  At this stage, the static and feature maps that were processed independently in previous stages are concatenated along their channel dimensions.


\subsection{Training Details}

Due to the use of an image decoder, rays must be processed in patches rather than sampled randomly throughout the image.  In the case of using only one or two tier scales, the entire image can be used in one training pass as compared to a subset of image pixels in previous NeRF methods.  This benefits the quality of the final rendering and contributes to the large decrease in rendering times.

In order to get estimated camera poses, we use pix-sfm \cite{lindenberger2021pixel} with feature type \textit{superpoint}, matcher type \textit{superglue} and sfm tool \textit{hloc}.  This is applied to all images from each data subset before splitting into training and validation.  For calculating dynamic PSNR, bounding boxes are needed for the dynamic regions of a scene, e.g. the people.  Okutama-Action has ground truth bounding boxes, however UG2 does not.  In order to rectify this, we used YOLOv8 to extract bounding boxes for each person in the UG2 video.

We use a NVIDIA RTX 3090 GPU for training with a learning rate of $1e-3$ for both the feature grids as well as the image decoder.  The training is run for 100,000 iterations using an entire image per training iteration.  Hyperparameter details between each method are kept as close as possible so as to offer a fair comparison between each method.  Although 4D Gaussian Splatting uses a different mechanic than NeRF, it does use a similar grid system with multiple scales as K-Planes.  The feature vector sizes within these grids are set so as to equal that of two tiers of the feature vectors in TK-Planes.

\section{Results}
\subsection{Datasets} 

\subsubsection{\textbf{Okutama-Action Dataset}}
The Okutama-Action dataset \cite{barekatain2017okutama} is collected using unmanned aerial vehicles (UAVs). It comprises scenes captured at various altitudes, angles, and time periods, containing different humans in 12 different action classes.  This dataset presents challenges due to dynamic transitions between actions, humans of multiple concurrent actions, and actors labeled with multiple actions. We train on the three different scenes: 1.1.1, 1.2.2, 1.2.6.  Each of these scenes offers distinct views of a baseball diamond at different times of day with varying numbers of people present, see Figures \ref{subfig:okutama-1.1.1}, \ref{subfig:okutama-1.2.2}, \ref{subfig:okutama-1.2.6}.
\subsubsection{\textbf{UG2 Dataset}}
\begin{figure}[t]
\vspace{2mm}
    \centering
    \subfigure[Okutama: 1.1.1]{\includegraphics[width=0.23\textwidth]{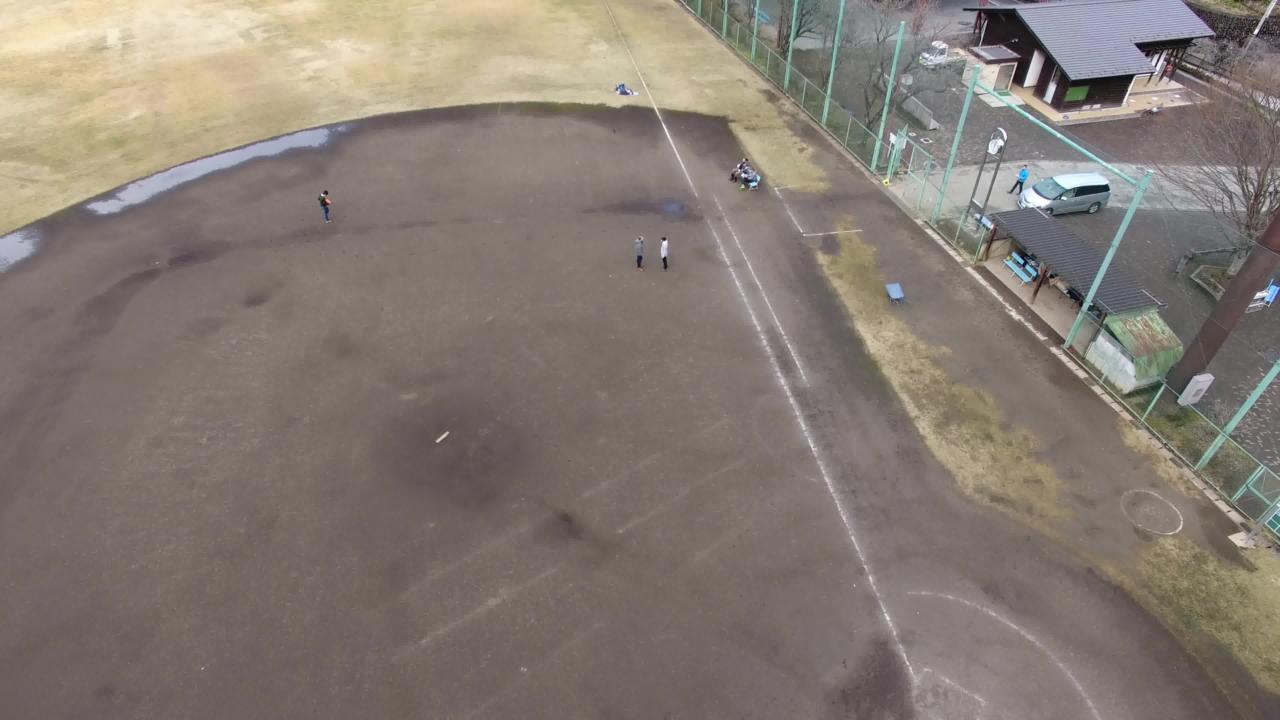}\label{subfig:okutama-1.1.1}}
     \subfigure[Okutama: 1.2.2]{\includegraphics[width=0.23\textwidth]{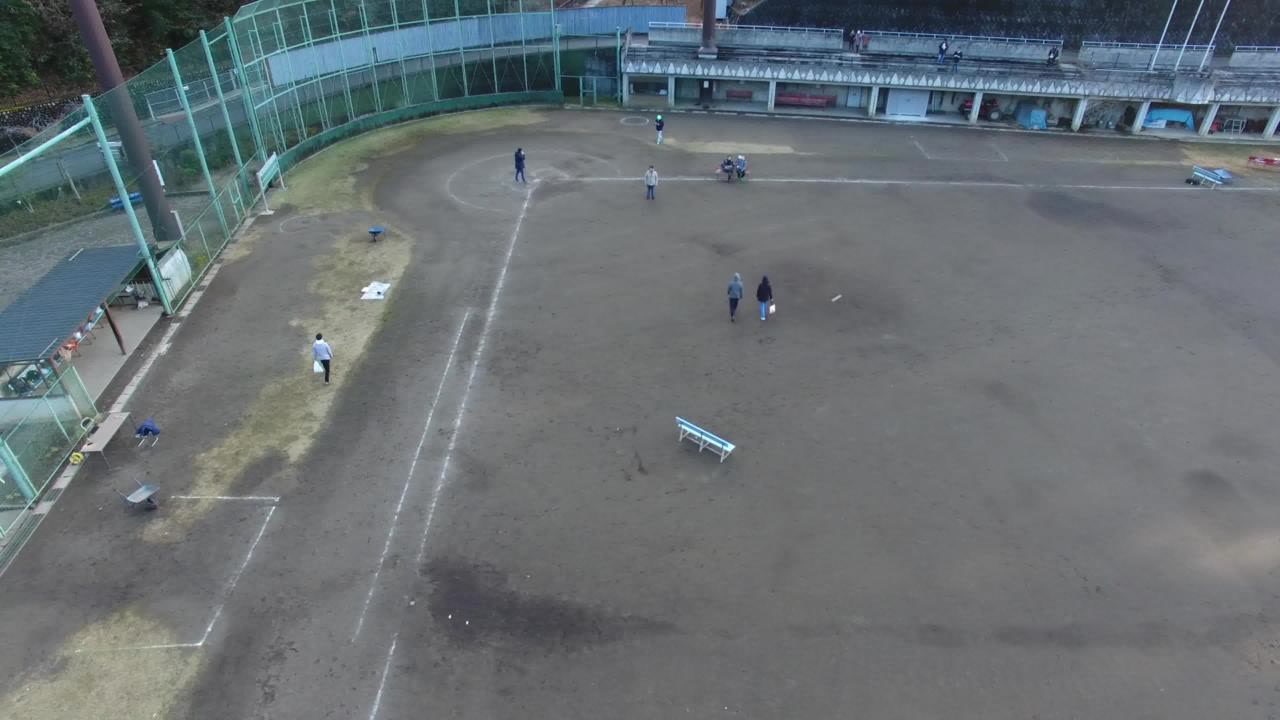}\label{subfig:okutama-1.2.2}}
     \subfigure[Okutama: 1.2.6]{\includegraphics[width=0.23\textwidth]{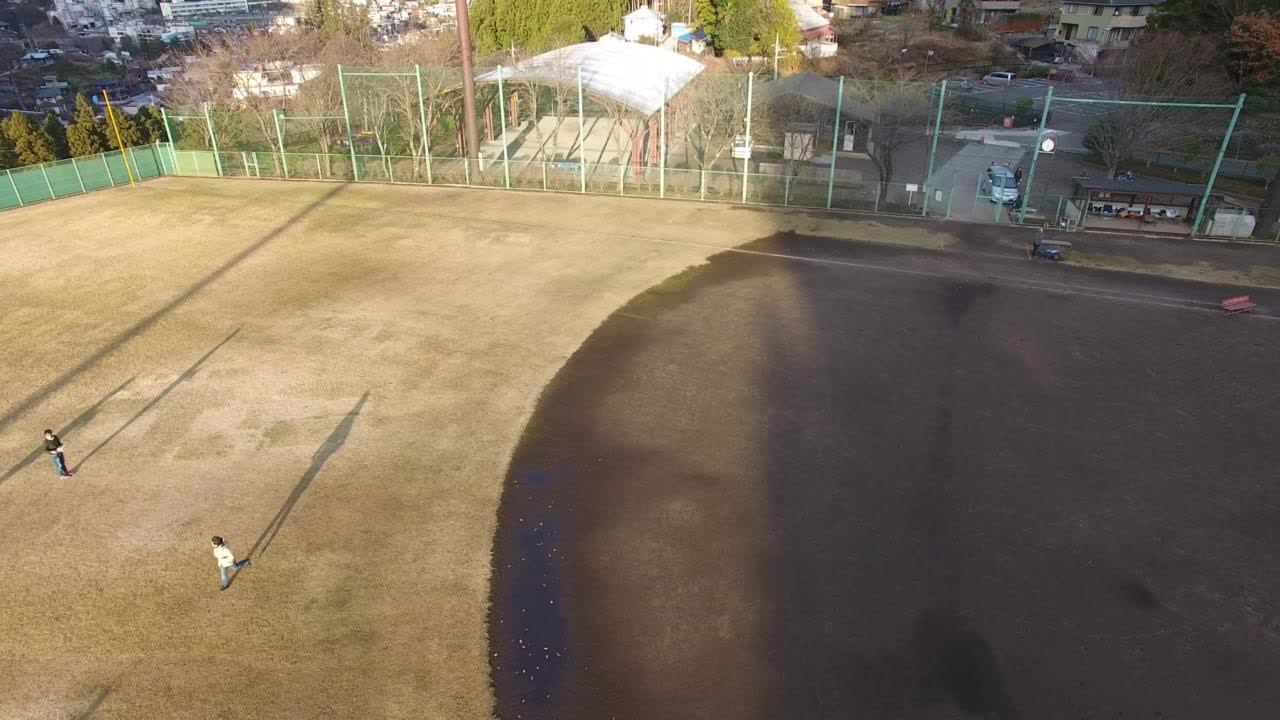}\label{subfig:okutama-1.2.6}}
      \subfigure[UG2: UAV-Test-133]{\includegraphics[width=0.23\textwidth]{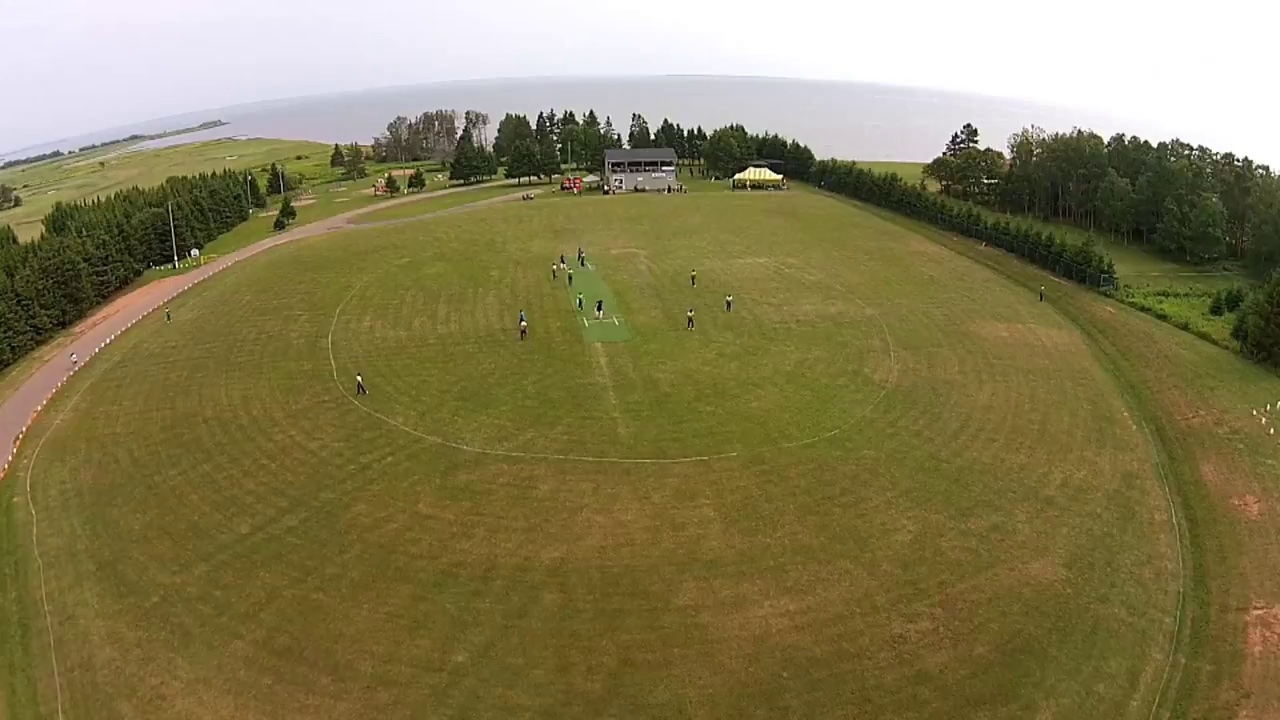}\label{subfig:ug2-uavtest-133}}    
    \caption{A example frame from each of the four scenes used to validate our experiments.  Each scene from Okutama-Action offers a unique viewpoint, time-of-day and number of people.  The scene from UG2 offers a challenging viewpoint from a quadrocopter of a cricket match in session.}
    \label{fig:dataset_examples}
\end{figure}

We also sample a video from the UG2 dataset \cite{vidal2018ug} which consists of a large variety of videos from varying styles of aerial vehicles, including piloted aircraft and UAVs such as fixed wing gliders and quadrocopters.  Although many of the videos within UG2 are not applicable for this given application, video 133 within the test set for UAV videos offers a challenging view of a cricket match in progress.  This particular video contrasts well with the scenes from Okutama-Action, offering smaller and more dynamic targets moving with less precise and scripted actions \ref{subfig:ug2-uavtest-133}.


\begin{table}[t]
\centering
\caption{PSNR Comparison on Okutama and UG2 Validation Subsets}
\label{table_psnr}
\resizebox{0.99\linewidth}{!}{
\begin{tabular}{c|c|c|c|c}
    \toprule
    Okutama  & K-Planes \cite{fridovich2023k} & Ext. K-Planes \cite{maxey2023uav} & 4D Gaussian \cite{wu20244d} & TK-Planes (Ours) \\
    \midrule
    \normalsize{\textbf{1.1.1}} & \normalsize{18.23} & \normalsize{32.61}  & \normalsize{33.95} & \normalsize{\textbf{34.41}} \\
    \normalsize{\textbf{1.2.2}} & \normalsize{14.45} & \normalsize{32.63}  & \normalsize{\textbf{34.45}} & \normalsize{34.40} \\
    \normalsize{\textbf{1.2.6}} & \normalsize{14.37} & \normalsize{27.89}  & \normalsize{29.35} & \normalsize{\textbf{30.28}} \\
    \midrule
    UG2  & K-Planes \cite{fridovich2023k} & Ext. K-Planes \cite{maxey2023uav} & 4D Gaussian \cite{wu20244d} & TK-Planes (Ours) \\
    \midrule
    \textbf{UAV-133} & \normalsize{17.11} & \normalsize{27.56} & \normalsize{30.47} & \normalsize{\textbf{31.52}} \\
    \bottomrule
\end{tabular}}
\end{table}

\begin{figure}[ht]
\vspace{2mm}
    \centering
    \subfigure[]{%
        \includegraphics[width=0.073\textwidth]{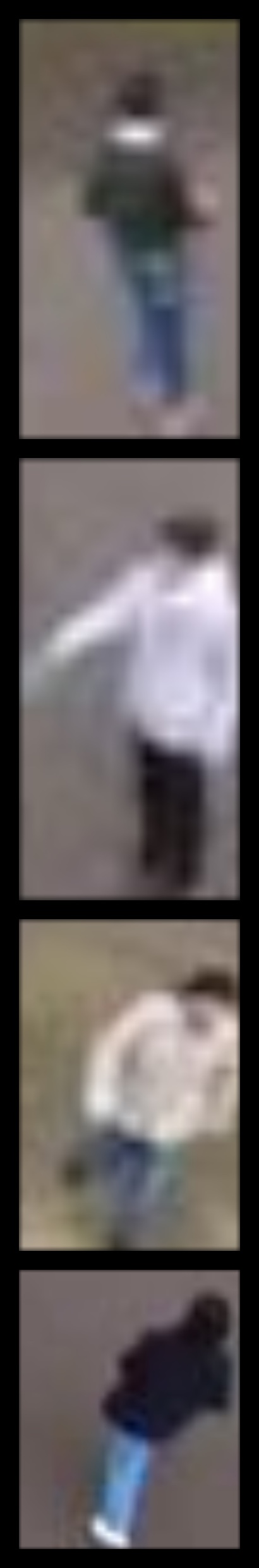}
        \label{subfig:compare_okutama_exp1}
    }
    \subfigure[]{%
        \includegraphics[width=0.073\textwidth]{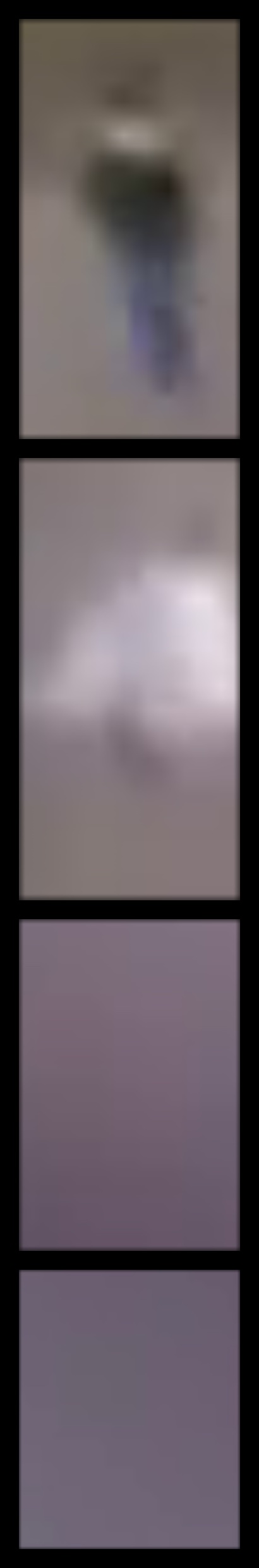}
        \label{subfig:compare_kplanes_exp1}
    }
    \subfigure[]{%
        \includegraphics[width=0.073\textwidth]{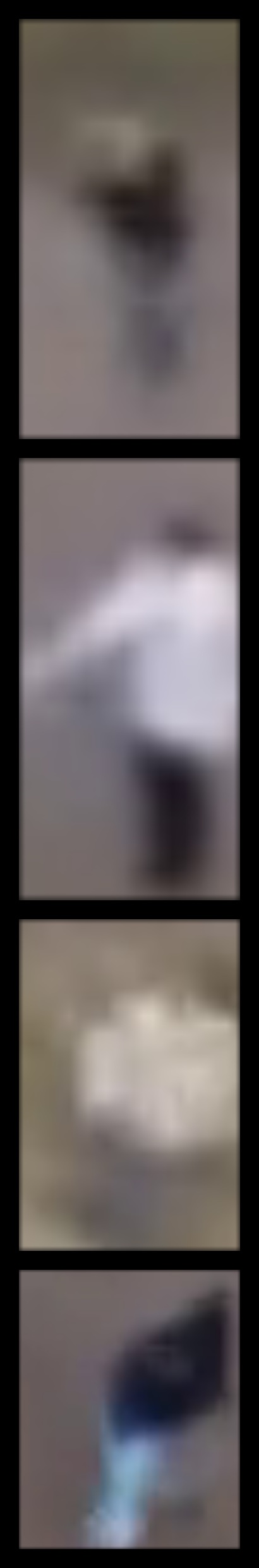}
        \label{subfig:compare_extended_exp1}
    }
    \subfigure[]{%
        \includegraphics[width=0.073\textwidth]{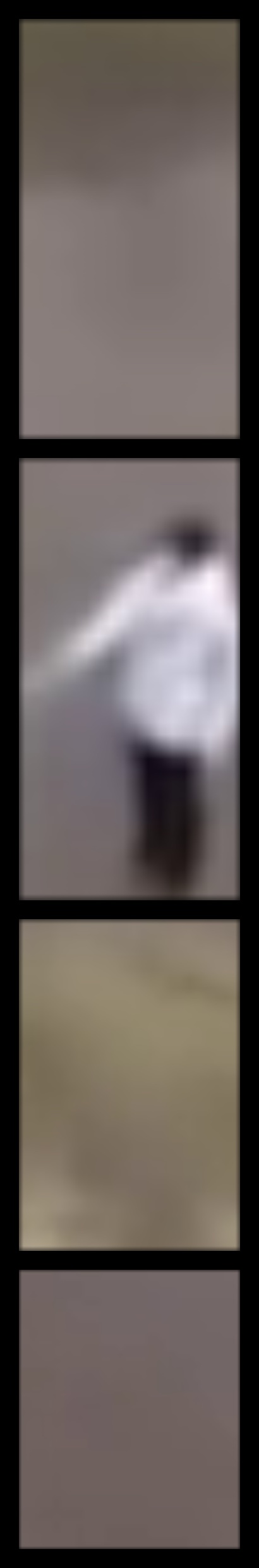}
        \label{subfig:compare_tk_exp1}
    }
    \subfigure[]{%
        \includegraphics[width=0.073\textwidth]{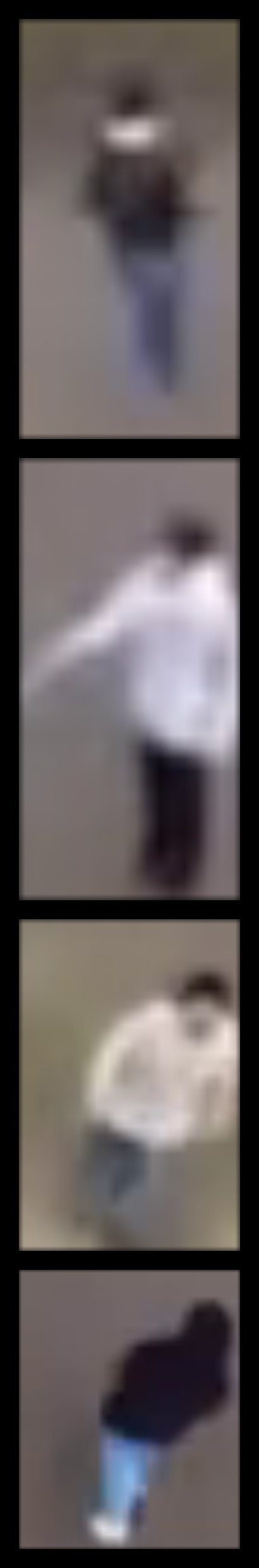}
        \label{subfig:compare_4DG_exp1}
    }
    \vspace{-2mm}
    \caption{A comparison of dynamic regions from a subsection of video 1.1.1 in Okutama-Action. 
    a.) Okutama ground truth, b.) stock K-Planes, c.) Extended K-Planes, d.) 4D-Gaussian, e.) TK-Planes. 
    We highlight the improved rendering quality generated by TK-Planes over the other methods on these challenging dynamic scenes.}
    \label{fig:dynamic_compare_exp1}
\end{figure}


\begin{table}[t]
\centering
\caption{Dynamic PSNR Comparison on Okutama-Action and UG2 Validation Subsets}
\label{table_dynamic_psnr}
\resizebox{0.99\linewidth}{!}{
\begin{tabular}{c|c|c|c|c}
    \toprule
    Okutama  & K-Planes \cite{fridovich2023k} & Ext. K-Planes \cite{maxey2023uav} & 4D Gaussian \cite{wu20244d} & TK-Planes (Ours) \\
    \midrule
    \normalsize{\textbf{1.1.1}} & \normalsize{20.55} & \normalsize{24.07} & \normalsize{24.56} & \normalsize{\textbf{28.17}}  \\
    \normalsize{\textbf{1.2.2}} & \normalsize{15.63} & \normalsize{24.52}  & \normalsize{24.69} & \normalsize{\textbf{27.97}} \\
    \normalsize{\textbf{1.2.6}} & \normalsize{16.31} & \normalsize{20.81}  & \normalsize{20.66} & \normalsize{\textbf{23.92}} \\
    \midrule
    Okutama  & K-Planes \cite{fridovich2023k} & Ext. K-Planes \cite{maxey2023uav} & 4D Gaussian \cite{wu20244d} & TK-Planes (Ours) \\
    \midrule
    \textbf{UAV-133} & \normalsize{15.54} & \normalsize{17.51}  & \normalsize{16.44} & \normalsize{\textbf{21.02}} \\
    \bottomrule
\end{tabular}}
\end{table}

\begin{figure}[ht]
    \vspace{2mm}
    \centering
    \subfigure[]{%
        \includegraphics[width=0.073\textwidth]{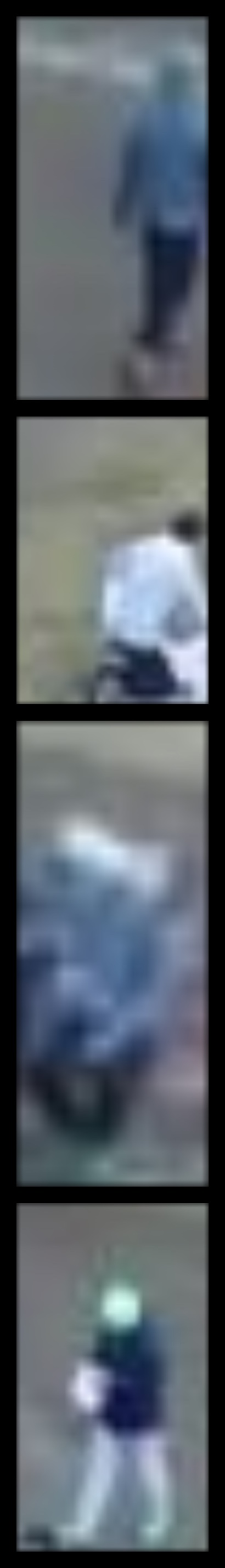}
        \label{subfig:compare_okutama_exp2}
    }
    \subfigure[]{%
        \includegraphics[width=0.073\textwidth]{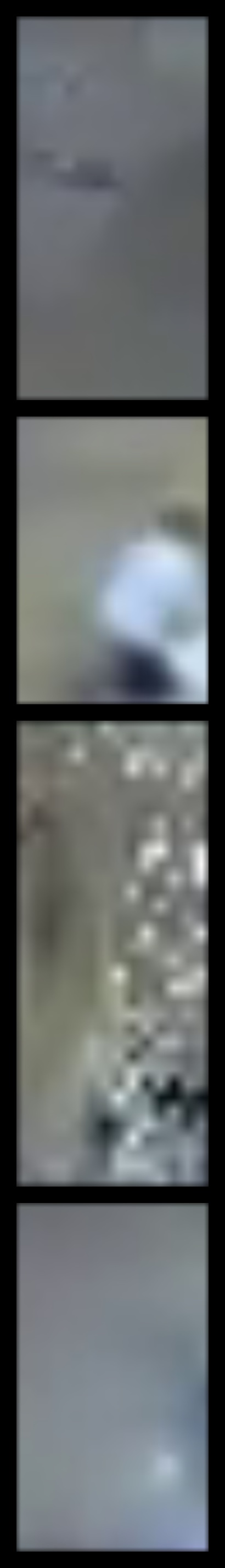}
        \label{subfig:compare_kplanes_exp2}
    }
    \subfigure[]{%
        \includegraphics[width=0.073\textwidth]{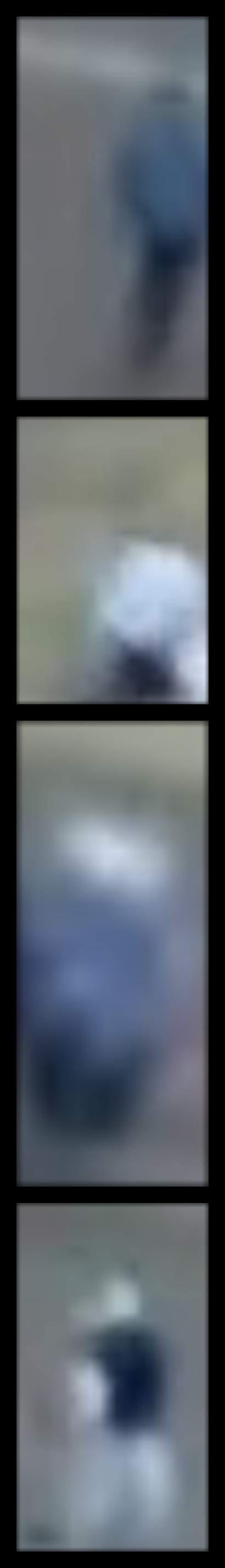}
        \label{subfig:compare_extended_exp2}
    }
    \subfigure[]{%
        \includegraphics[width=0.073\textwidth]{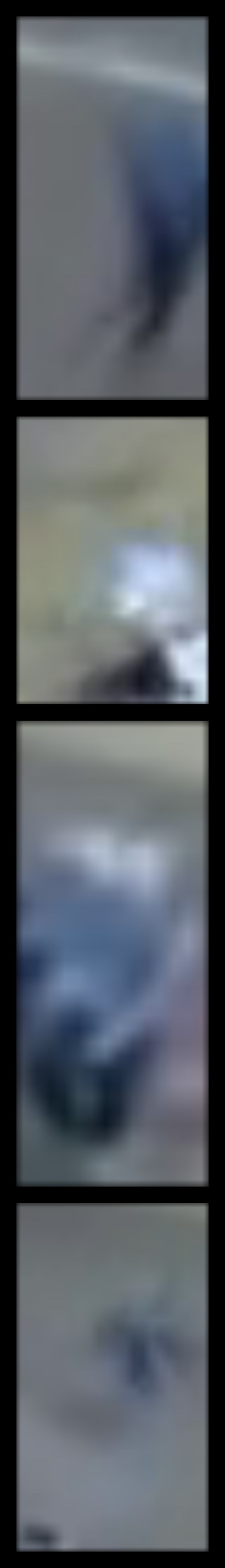}
        \label{subfig:compare_tk_exp2}
    }
    \subfigure[]{%
        \includegraphics[width=0.073\textwidth]{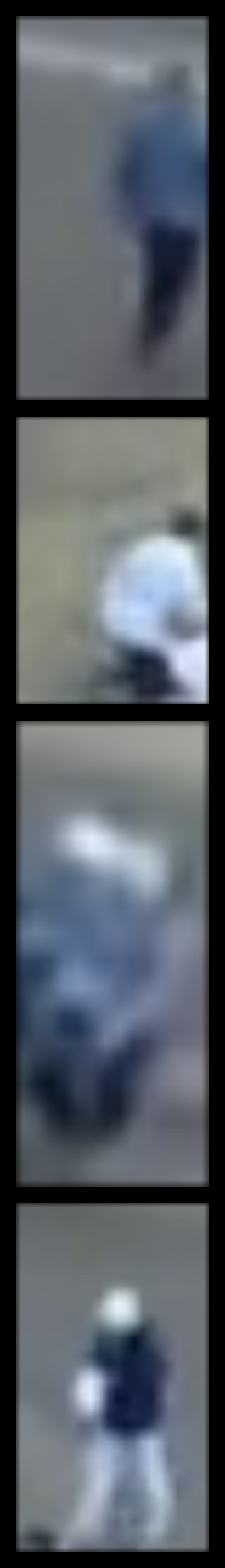}
        \label{subfig:compare_4DG_exp2}
    }
    \vspace{-2mm}
    \caption{A comparison of dynamic regions from a subsection of video 1.2.2 in Okutama-Action. 
    a.) Okutama ground truth, b.) stock K-Planes, c.) Extended K-Planes, d.) 4D-Gaussian, e.) TK-Planes. 
    We highlight the improved rendering quality generated by TK-Planes over the other methods on these challenging dynamic scenes.}
    \vspace{-3mm}
    \label{fig:dynamic_compare_exp2}
\end{figure}

\begin{figure}[h!t]
    \vspace{2mm}
    \centering
    \subfigure[]{%
        \includegraphics[width=0.073\textwidth]{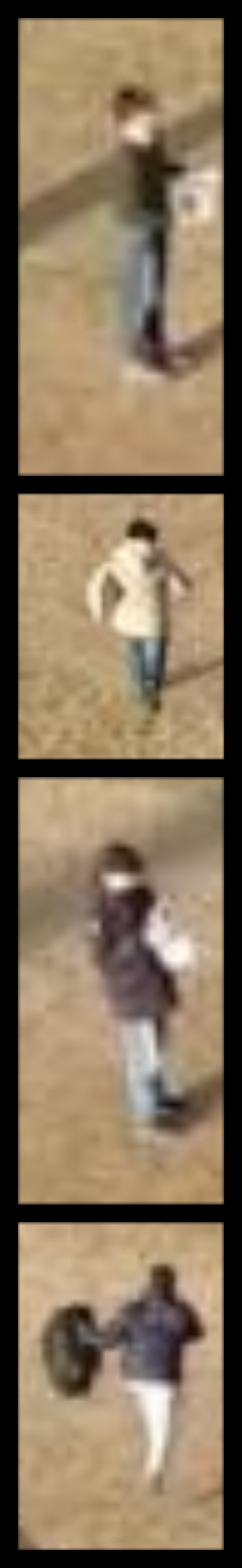}
        \label{subfig:compare_okutama_exp3}
    }
    \subfigure[]{%
        \includegraphics[width=0.073\textwidth]{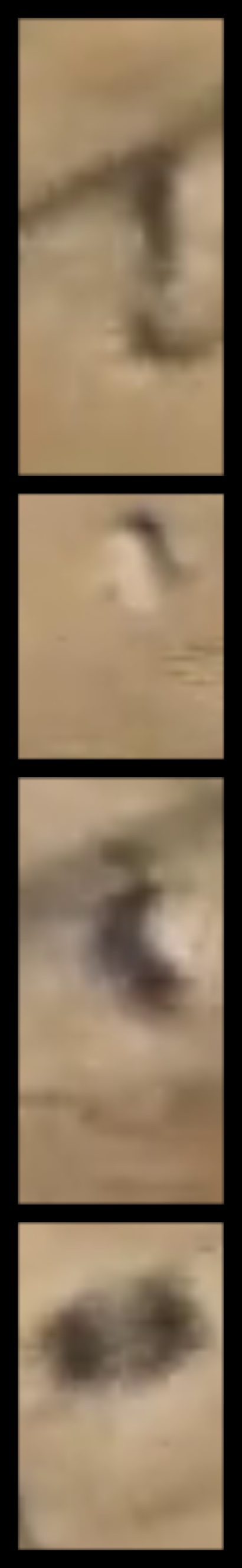}
        \label{subfig:compare_kplanes_exp3}
    }
    \subfigure[]{%
        \includegraphics[width=0.073\textwidth]{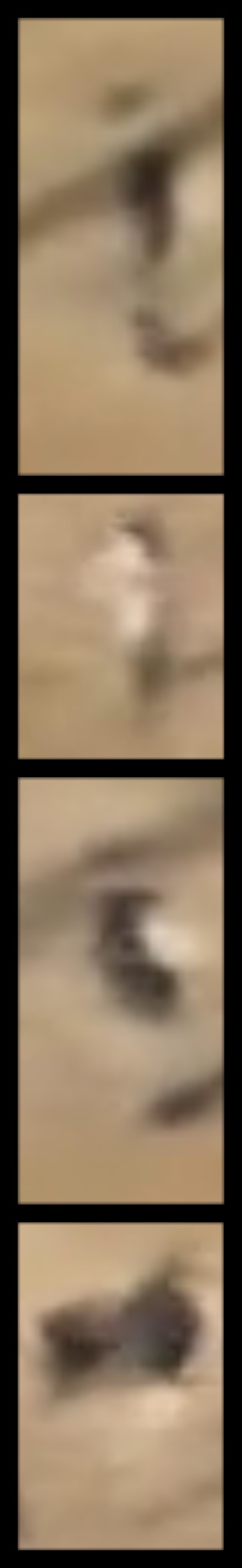}
        \label{subfig:compare_extended_exp3}
    }
    \subfigure[]{%
        \includegraphics[width=0.073\textwidth]{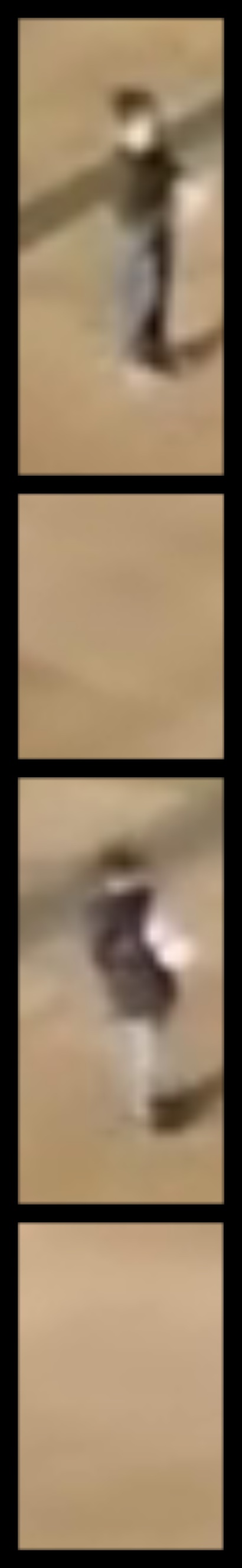}
        \label{subfig:compare_tk_exp3}
    }
    \subfigure[]{%
        \includegraphics[width=0.073\textwidth]{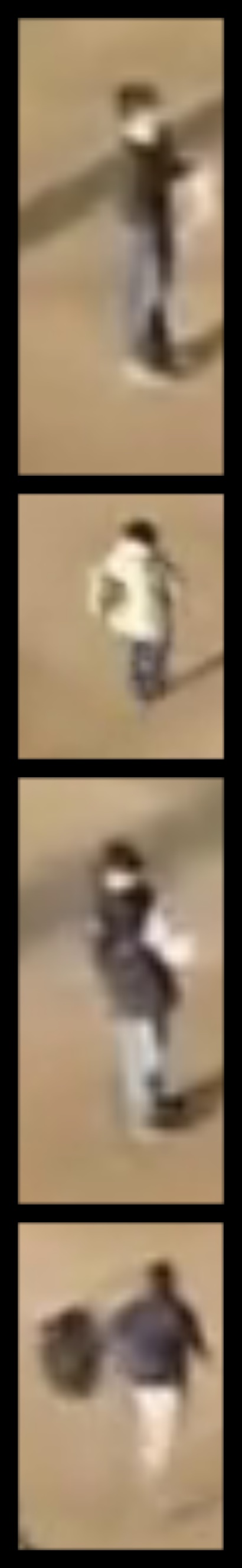}
        \label{subfig:compare_4DG_exp3}
    }
    \vspace{-2mm}
    \caption{A comparison of dynamic regions from a subsection of video 1.2.6 in Okutama-Action. 
    a.) Okutama ground truth, b.) stock K-Planes, c.) Extended K-Planes, d.) 4D-Gaussian, e.) TK-Planes. 
    We highlight the improved rendering quality generated by TK-Planes over the other methods on these challenging dynamic scenes.}
    \vspace{-5mm}
    \label{fig:dynamic_compare_exp3}
\end{figure}

\begin{figure}[ht]
    \vspace{2mm}
    \centering
    \subfigure[]{%
        \includegraphics[width=0.073\textwidth]{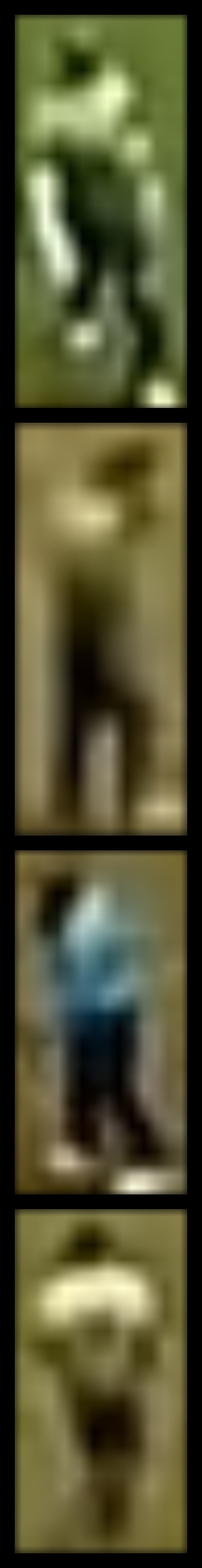}
        \label{subfig:compare_ug2_exp5}
    }
    \subfigure[]{%
        \includegraphics[width=0.073\textwidth]{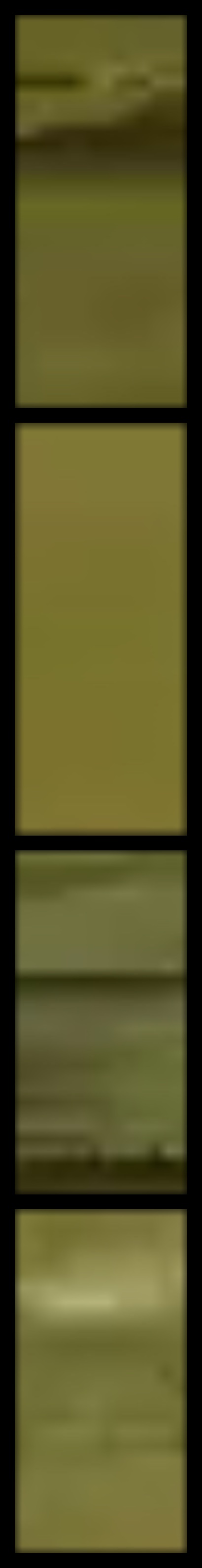}
        \label{subfig:compare_kplanes_exp5}
    }
    \subfigure[]{%
        \includegraphics[width=0.073\textwidth]{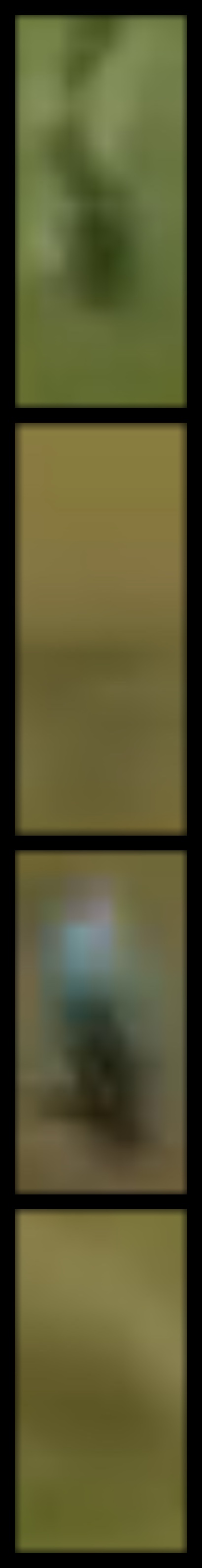}
        \label{subfig:compare_extended_exp5}
    }
    \subfigure[]{%
        \includegraphics[width=0.073\textwidth]{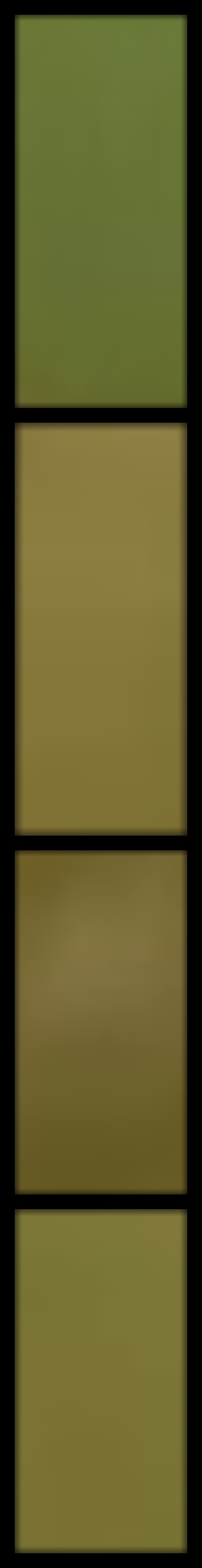}
        \label{subfig:compare_tk_exp5}
    }
    \subfigure[]{%
        \includegraphics[width=0.073\textwidth]{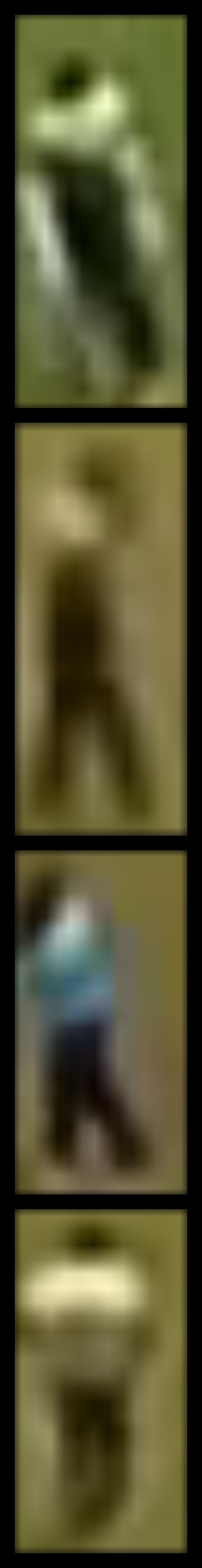}
        \label{subfig:compare_4DG_exp5}
    }
    \vspace{-2mm}
    \caption{A comparison of dynamic regions from a subsection of video UAV-133 in UG2. 
    a.) UG2 ground truth, b.) stock K-Planes, c.) Extended K-Planes, d.) 4D-Gaussian, e.) TK-Planes. 
    The UG2 dataset is particularly challenging; however, it can be seen that TK-Planes performs admirably compared to the other methods.}
    \vspace{-5mm}
    \label{fig:dynamic_compare_exp5}
\end{figure}


\subsection{Analysis}

Because we are focused on dynamic portions of a scene, we calculate not only overall PSNR values but also what we call Dynamic-PSNR (DPSNR) values which represent the average PSNR of each bounding box for people within each frame.  With respect to tasks such as object detection, pose recognition and action recognition, the fidelity of the target object is key and thus the most pertinent for validation.  As can be seen from Tables \ref{table_psnr} and \ref{table_dynamic_psnr}, the overall performance of TK-Planes outperforms stock and extended K-Planes in all experiments, up to 1.8 PSNR in Okutama tests and 3.96 PSNR in the UG2 test.  With respect to DPSNR, TK-Planes takes a wider lead with up to 4.1 DPSNR improvement in Okutama tests and 3.51 DPSNR in the very difficult UG2 test.  

Furthermore, TK-Planes with two tiers of feature vector grids runs 10x faster when rendering than previous versions of K-Planes.  A one tier version of TK-Planes still outperforms previous K-Planes methods while running 25x faster when rendering. This is due to the lower number of rays and sampling points along each ray required by TK-Planes versus K-Planes or other previous state of the art NeRF methods.  It is noted that the version of TK-Planes in this paper uses the simple ray sampling method of uniform sampling, compared to the more advanced proposal sampling method of K-Planes and extended K-Planes.  See \cite{maxey2024tk} for more comparisons between one and two tier TK-Planes.

4D Gaussian Splatting \cite{wu20244d} is a state of the art neural rendering method that outperforms many previous state of the art NeRF methods, including K-Planes.  With respect to UAV videos with estimated camera poses, TK-Planes outperforms 4D Gaussians in the Okutama tests by up to 0.93 PSNR, and 1.05 PSNR on the difficult UG2 test.  The performance gap is widened with respect to DPSNR, with up to 3.61 DPSNR improvement in the Okutama tests and 4.58 DPSNR in the UG2 test.  X\ref{fig:dynamic_compare_exp1}, \ref{fig:dynamic_compare_exp2}, \ref{fig:dynamic_compare_exp3} and \ref{fig:dynamic_compare_exp5} showcase some examples of DSPNR regions, e.g. parts of the frame with people.  The performance improvement for dynamic people can be qualitatively seen in these images, in particular for the UG2 test, in Figure \ref{fig:dynamic_compare_exp5}. See \cite{maxey2024tk} for more comparisons between K-Planes, extended K-Planes, TK-Planes and 4D Gaussians.

\section{Conclusion and Future Work}

We have presented a novel method for synthetic data generation. Our tiered K-Planes algorithm has shown that it can improve the fidelity of dynamic objects within difficult to render scenes such as those from UAVs.  In certain cases this improvement is upwards of 3.96 PSNR and 4.1 DPSNR over previous NeRF methods and 1.05 PSNR and 4.58 DPSNR over a state of the art 4D Gaussian Splatting method.  Overall, the increase in fidelity of dynamic objects will help with tasks such as object detection, pose recognition and action recognition wherein the details of the subjects are paramount.

Future work with respect to TK-Planes involves a more robust grid search of hyperparameters given particular scenes and environmental conditions.  Furthermore, the use of abstract feature vectors can be exploited to take advantage of redundant information within a scene and improve to a greater degree the fidelity of images and novelty of poses and time-slices of dynamic scenes.

\bibliographystyle{plain}
\bibliography{References}

\end{document}